\documentclass{article}

\PassOptionsToPackage{numbers,compress}{natbib}
\usepackage[final]{tackling_climate_workshop_style}

\usepackage[utf8]{inputenc}
\usepackage[T1]{fontenc}
\usepackage{hyperref}
\hypersetup{hidelinks}
\usepackage{url}
\usepackage{booktabs}
\usepackage{amsfonts}
\usepackage{nicefrac}
\usepackage{microtype}
\usepackage{graphicx}
\usepackage{amsmath}
\usepackage{amssymb}
\usepackage{float}

\newcommand{\nPop}{15{,}763}            %
\newcommand{\nZones}{15}
\newcommand{\nRoads}{8}
\newcommand{\nShelters}{3}
\newcommand{\nDepots}{2}
\newcommand{\nHazardNodes}{2}

\newcommand{\depthExposed}{20}        %
\newcommand{\depthRoad}{30}           %
\newcommand{\sandbagDelta}{40}        %

\newcommand{\nTuneWindows}{16}
\newcommand{\nTestWindows}{18}

\newcommand{\nTrainSteps}{500{,}000}
\newcommand{\nEnvs}{8}
\newcommand{\nSeedsPPO}{5}            %
\newcommand{\wdDefault}{0.2}
\newcommand{\wdLow}{0.1}
\newcommand{\wdHigh}{0.4}
\newcommand{\wCost}{0.05}

\newcommand{\nGridStatic}{300}

\newcommand{\tauStatic}{120}

\newcommand{\harmMonitor}{2,424}
\newcommand{\dispMonitor}{40,920}
\newcommand{\peakMonitor}{805}
\newcommand{\succMonitor}{0.39}

\newcommand{\harmPCEx}{1,439}
\newcommand{\dispPCEx}{47,269}
\newcommand{\peakPCEx}{722}
\newcommand{\succPCEx}{0.39}

\newcommand{\harmStatic}{308}
\newcommand{\dispStatic}{41,721}
\newcommand{\peakStatic}{233}
\newcommand{\succStatic}{0.72}

\newcommand{\harmPPO}{262}
\newcommand{\harmPPOstd}{26}
\newcommand{\dispPPO}{42,007}

\newcommand{\peakPPO}{187}
\newcommand{\succPPO}{0.77}

\newcommand{\ciLo}{$-172$}               %
\newcommand{\ciHi}{$+34$}

\newcommand{\harmNoMsg}{285}
\newcommand{\harmNoMsgStd}{77}
\newcommand{\dispNoMsg}{42,007}
\newcommand{\peakNoMsg}{210}
\newcommand{\succNoMsg}{0.74}

\newcommand{\shareLowestZone}{90\%}

\newcommand{\harmStaticB}{645}          %

\newcommand{\harmPPOB}{673}
\newcommand{\harmPPOBstd}{38}

\newcommand{\sigmaOne}{2}

\newcommand{\sigmaTwelve}{6}

\newcommand{\nTiedWindows}{16}                %

\title{GNN-Based Deep Reinforcement Learning for Coastal Emergency Response under Extreme Water Levels}

\author{%
  Marcello Sano$^{1,2,3,5}$, Samuele Casagrande$^{1}$, Davide Ferrario$^{1,2}$, Sebastiano Vascon$^{2,4}$, \\
  \textbf{Silvia Torresan}$^{1,2}$, \textbf{Andrea Critto}$^{2,1}$ \\[2pt]
  $^{1}$CMCC Foundation -- Euro-Mediterranean Center on Climate Change, Venice, Italy \\
  $^{2}$Department of Environmental Sciences, Informatics and Statistics, \\ Ca' Foscari University of Venice, Italy \\
  $^{3}$Griffith University, Gold Coast, Australia \\
  $^{4}$European Centre for Living Technology, Venice, Italy \quad $^{5}$Apter, Byron Bay, Australia
}

\begin{document}

\maketitle

\begin{abstract}
When a storm surge approaches a coastal town, emergency response depends on how quickly and effectively an operational plan can be implemented using the available information. This may include real-time water-level forecasts, expected flood extent, road access, the location of vulnerable residents, shelters and available resources. Bringing this information together under time pressure can be slow and depends on the duty officer's expertise, potentially delaying critical decisions when they matter most. In this research, we ask whether a policy can learn the timing of emergency actions from decades of real events using a spatial representation of the territory. Lido di Venezia is represented as a heterogeneous graph built from public elevation, road, building and census data and driven by hourly tide-gauge records from 1983 to 2024. Seven possible emergency actions are derived from the municipal emergency plan. We train a PPO decision policy using a graph representation that links residential areas, roads, shelters and hazard zones. The trained policy is then tested on storm windows not used during training and compared with a rule-based system simulating the officer's discretionary decisions. On \nTestWindows{} historical test scenarios spanning low to severe conditions, PPO achieves a similar reduction in population exposure to the rule-based system. In a second test, with changing and uncertain forecasts and limited protection, PPO again performs similarly but does not outperform the tuned rule. In a third test, removing graph message passing does not substantially reduce performance. Together, these tests show that PPO can learn an effective protective response from an outcome-based reward alone: it is given the plan's actions and the conditions under which each is allowed, but not the thresholds that tell the comparator when to act. Future research will test the learned policy on more complex edge cases and replace the simulated officer with direct assessment by emergency-management stakeholders, supported by large language model (LLM) based interpretation of the recommendations.
\end{abstract}

\section{Introduction}
\label{sec:intro}

When a storm surge approaches Lido di Venezia, the duty officers are responsible to implement an Emergency Plan \citep{comunevenezia2008piano} by observing data feeds, including water level forecasts, streets elevation, where vulnerable residents live, which shelters are open, among other information. This plan has two layers. Two actions are mandatory and executed directly on numeric triggers, an alert at $\geq 110$\,cm and area evacuation at $\geq 150$\,cm. Five actions are discretionary, including sandbags ``at exposed points'', assisted evacuation where a route is passable, shelters, road closures and continued surveillance (Table~\ref{tab:actions} for details), whose choice and timing are left to human judgment. Bringing the data feed and the plan together under time pressure can be slow and depends on local expertise limited by the operator's experience.

In our research, we explore if a policy can learn the timing of the discretionary actions from the record of real events, given a representation of the territory that carries its geography. We compare the performance of the policy with a simple, rule based simulated human decision based on the idea ``if the $h$-hour forecast maximum reaches $\tau$, do action $X$, at most $q$ times'', which is based on the ideal world where an officer knows exactly what to do and when based on the rules given, which is often not the case. On the other hand, the learned policy is given only the objective, save lives, minimise displacement and cost, and never the rules. Importantly, the mandatory actions in the plan are not used for training, which focuses on the 5 actions that need human judgement.

\textbf{Contributions.} \emph{(i)} An event-scale environment on real geography: the Lido graph from public data, seven plan-derived actions, and hourly replay of the 1983--2024 tide-gauge record. \emph{(ii)} An evaluation protocol: rules represent the simulated officer, with tuning and test events and a pass/fail criterion fixed in advance. \emph{(iii)} Three tests: in Test~1, with accurate forecasts and flexible protection, PPO reduces population exposure to a similar level as the tuned rule. In Test~2, with simulated changing and uncertain forecasts, PPO performs similarly but does not outperform the rule. In Test~3, removing graph message passing does not consistently reduce performance. The result is not that learning beats rules, but that PPO can uncover an effective response from outcomes alone, providing a basis for testing harder decisions where rules may be insufficient. Code and trained checkpoints are available at \url{https://github.com/marcellosano/coastal-emergency-drl}.

\subsection{Related work}

Reinforcement learning has already been applied to coastal and flood risk, mainly in two ways: to support long-term investment decisions under sea-level-rise uncertainty \citep{feng2025rlcoastal,shuvo2022multiagentslr,costa2024climaterl}, and to optimise evacuation during individual or synthetic hazard events \citep{tang2024transitevac,viola2024evacrl}. Graph neural networks have also been used for flood modelling and infrastructure control \citep{bentivoglio2023swegnn,meirom2021graphcontrol,yoon2021l2rpn}, while action masking is commonly used to prevent an agent from selecting actions that are not allowed \citep{huang2020closer,kanervisto2020actionspace}. In practice, however, emergency preparedness still relies heavily on threshold-based rules and triggers \citep{gomezrave2025preparedness,ferrarin2020adriaticstorm}. Our work brings these elements together in a different way: hourly emergency decisions over a representation of real geography, actions and constraints derived from an operational plan, four decades of replayed water-level events, and a rule-based version of the plan's judgement layer as the main comparator.

\section{Environment and policy}
\label{sec:env}

\begin{figure}[b]
  \centering
  \includegraphics[width=0.72\linewidth]{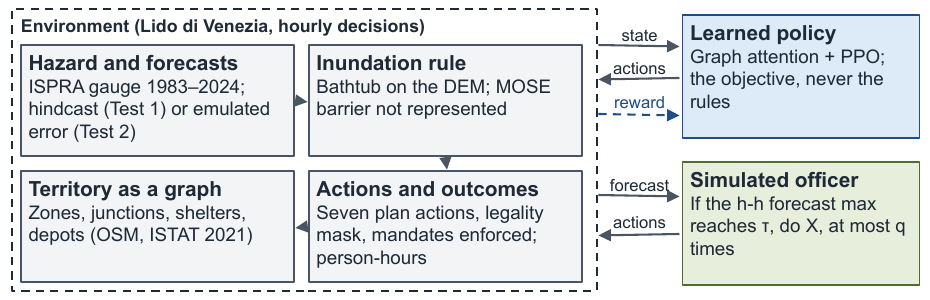}
  \caption{The environment and the two deciders, connected by state, actions and reward.}
  \label{fig:framework}
\end{figure}

Figure~\ref{fig:framework} shows the environment and the two deciders that act on it.

\subsection{Hazard, forecasts and inundation}
\label{sec:hazard}

For the hazard, we replay the ISPRA Diga Sud Lido hourly record 1983--2024. The forecast is the gauge value 1\,h ahead and its maxima over 6 and 12\,h. In Test~1 the forecast is a hindcast (perfect) and protection can stack. In Test~2 an emulator adds a forecast error growing with lead time (\sigmaOne{}--\sigmaTwelve{}\,cm over 1--12\,h) and protection is capped at two sandbag deployments. For inundation, we use a bathtub rule: the gauge level, shifted to the elevation datum, is compared with ground height plus local protection; a zone is exposed at a modelled depth $\geq \depthExposed{}$\,cm and a junction impassable at $\geq \depthRoad{}$\,cm. For simplicity, the rule ignores the MOSE barrier, so post-2020 lagoon-side inundation is counterfactual \citep{lionello2021venicereview}.

\subsection{The territory as a graph}
\label{sec:graph}

For the territory, we build a graph from the gauge record, the ISTAT 2021 census, the Tinitaly elevation model \citep{tinitaly}, OpenStreetMap \citep{osm} and the plan: \nZones{} residential zones, \nRoads{} junctions, \nShelters{} shelters, \nDepots{} depots and \nHazardNodes{} hazard nodes. We chose a graph because it represents the geography and is the path toward a digital twin: messages along edges give each zone the state of its junctions and shelters.

\subsection{Plan-derived actions and outcomes}
\label{sec:actions}

\textbf{Actions:} The plan gives the decider seven possible actions (Table~\ref{tab:actions}). Each one is allowed only under certain conditions, takes up to an hour to have an effect, and uses part of a small crew budget. Two of the seven are mandatory: the alert and the area evacuation. The environment triggers these itself when the water-level thresholds are reached, for every decider alike, so they are never learned. The comparison between the learned policy and the rules therefore concerns only the five discretionary actions. A decider chooses \emph{what} to do and \emph{when}, but not \emph{where}: if it chooses to deploy sandbags, the environment sends them to the zone with the highest forecast flood depth. At this stage, no decider makes a spatial choice.

\textbf{Outcomes:} Every hour we count three things: people living in flooded zones, people who were evacuated although their zone did not flood, and the crew cost of the action taken. Adding the first two up over a storm gives the two headline numbers of the paper, exposure person-hours and needless-displacement person-hours.

\begin{table}[t]
  \centering
  \caption{The seven plan-derived actions \citep{comunevenezia2008piano}: type, legality preconditions, effect, latency (h) and cost (budget 5).}
  \label{tab:actions}
  \scriptsize
  \setlength{\tabcolsep}{2.5pt}\renewcommand{\arraystretch}{0.92}
  \begin{tabular}{@{}l l l p{5.45cm} p{2.3cm} c c@{}}
    \toprule
    Action & Plan \S & Type & Preconditions & Effect & Lat. & Cost \\
    \midrule
    monitor              & 4.1.4.1, 4.1.5.5 & default          & always legal (no-op) & --- & 0 & 0 \\
    issue\_alert         & 4.1.4.9          & \textbf{mandate} & 1-h forecast $\geq 110$\,cm; not yet alerted & preparedness flag & 0 & 0 \\
    deploy\_sandbags     & 4.1.4.8          & discr.\ & exposed zone; resources $\geq 1$ & $+\sandbagDelta{}$\,cm protection & 1 & 1 \\
    close\_road          & 4.1.4.9          & discr.\ & junction water level $\geq$ flood threshold & junction blocked & 0 & 0 \\
    open\_shelter        & 5.3.2, 4.1.4.8   & discr.\ & accessible; capacity $> 0$; not active; resources $\geq 1$ & shelter active & 1 & 1 \\
    assisted\_evacuation & 4.2.3.1, 4.2.3.4 & discr.\ & zone not evacuated; passable path to a shelter; resources $\geq 1$ & zone evacuated & 1 & 1 \\
    full\_evacuation     & 4.2.3.1, 4.1.5.7 & \textbf{mandate} & 1-h forecast $\geq 150$\,cm & all zones evacuated & 1 & 3 \\
    \bottomrule
  \end{tabular}
\end{table}

\subsection{Learning the policy}
\label{sec:method}

A graph attention encoder \citep{velickovic2018gat} with two message-passing layers yields node representations, mean-pooled per type and concatenated with the scalar state; a policy output scores the seven actions under the legality mask \citep{huang2020closer} and a value output estimates the return. The reward at hour $t$ is
\begin{equation}
  r_t = -\,\frac{\sum_i \mathbf{1}[\text{exposed}_{i,t}]\,\text{pop}_i}{P}
        \;-\; w_d\,\frac{\sum_i \mathbf{1}[\text{evacuated}_{i,t} \wedge \neg\text{exposed}_{i,t}]\,\text{pop}_i}{P}
        \;-\; w_r\,\frac{\text{cost}(a_t)}{\text{budget}} \,,
  \label{eq:reward}
\end{equation}
where $P = \nPop{}$ residents, $w_d = \wdDefault{}$ weighs needless displacement (\wdLow{} and \wdHigh{} were run as scenarios) and $w_r = \wCost{}$ weighs crew cost. We train with proximal policy optimisation (PPO) \citep{schulman2017ppo} for \nTrainSteps{} steps per seed on \nEnvs{} parallel environments, undiscounted over the window.

\section{Experiments and results}
\label{sec:experiments}

We evaluate on 24-hour storm windows, split by storm so that no tuning or test storm appears in training: \nTuneWindows{} windows tune the rule-based comparator and \nTestWindows{} held-out windows test every decider. We compare four ways of completing the plan. \emph{Mandates only} applies the two mandatory actions only. \emph{Plan-extended} adds the plan's own heuristic triggers for the discretionary actions. The \emph{static trigger} is the simulated officer of the Introduction: the best of \nGridStatic{} candidate rules, chosen on the tuning windows. The \emph{learned policy} is PPO, reported as mean $\pm$ std over \nSeedsPPO{} seeds; Test~3 repeats it with graph message passing switched off. The headline measure is exposure person-hours, compared window by window with the trigger. The criterion, fixed before any result was seen: the learned policy beats the rule only if its exposure is lower by more than the 95\% bootstrap interval on most seeds.

\begin{table}[t]
  \centering
  \caption{Test~1 ladder, \nTestWindows{} test windows (per-window means; learned rows mean $\pm$ std, \nSeedsPPO{} seeds).}
  \label{tab:ladder}
  \scriptsize
  \setlength{\tabcolsep}{4.5pt}\renewcommand{\arraystretch}{0.92}
  \begin{tabular}{@{}lrrrr@{}}
    \toprule
    Policy & Exposure (person-h) $\downarrow$ & Displacement (person-h) $\downarrow$ & Peak exposed (persons) & Zero-exposure windows \\
    \midrule
    Mandates only (literal plan)   & \harmMonitor{} & \dispMonitor{} & \peakMonitor{} & \succMonitor{} \\
    Plan-extended                  & \harmPCEx{}    & \dispPCEx{}    & \peakPCEx{}    & \succPCEx{}    \\
    Static trigger (tune-selected) & \harmStatic{}  & \dispStatic{}  & \peakStatic{}  & \succStatic{}  \\
    No message passing             & \harmNoMsg{} $\pm$ \harmNoMsgStd{} & \dispNoMsg{} & \peakNoMsg{} & \succNoMsg{} \\
    Learned policy                 & \harmPPO{} $\pm$ \harmPPOstd{} & \dispPPO{} & \peakPPO{} & \succPPO{} \\
    \bottomrule
  \end{tabular}
\end{table}

In Table~\ref{tab:ladder} (Test~1), the mandates alone leave \harmMonitor{} exposure person-hours per window. The tuned rule, one sandbag deployment when the 12-hour forecast reaches \tauStatic{}\,cm, brings this down to \harmStatic{}; the learned policy reaches \harmPPO{} $\pm$ \harmPPOstd{}. The difference is within the uncertainty (95\% interval [\ciLo{}, \ciHi{}]) and no seed meets the criterion. The learned policy matches the tuned rule and does not beat it. It found the same move as the simulated officer: on \nTiedWindows{} of \nTestWindows{} windows the outcomes are identical. In Test~3, removing message passing changes little (\harmNoMsg{} $\pm$ \harmNoMsgStd{}). In Test~2 (uncertain forecasts, capped protection), the rule tuned for those conditions does slightly better than the learned policy (\harmStaticB{} versus \harmPPOB{} $\pm$ \harmPPOBstd{}), and no seed passes.

This geography cannot produce cases beyond the rules: exposure is concentrated in three zones, one carrying \shareLowestZone{} of it, and the environment, not the decider, chooses where an action is applied, so each storm admits one good discretionary move and both deciders find it.

\section{Discussion and conclusions}
\label{sec:impact}

Sea-level rise will make extreme water-level events more frequent, increasing the need for emergency decisions that are timely, consistent and able to respond as conditions change. In this study, PPO broadly matches a rule-based system that represents a simplified version of the officer's discretionary decisions, despite not being given the decision thresholds used by that system. This suggests that the policy can learn an effective protective response directly from the outcomes defined in its reward function. However, the current environment remains relatively simple, and removing graph message passing does not substantially reduce performance because neither PPO nor the rule-based system currently chooses where actions are applied.

The next stage of the research will move beyond comparison with a simulated officer, testing the recommendations of the deep reinforcement learning (DRL) policy directly with emergency-management stakeholders across a wider range of scenarios, including edge cases with competing priorities, limited resources, changing forecasts and disrupted access. Practitioners may assess whether the recommendations are appropriate, useful and operationally credible \citep{dalbarco2025coastaid,dulacarnold2021realworld}. Operationally, we may run two heads side by side: a rule engine, i.e., the plan's judgement layer written as rules and auditable clause by clause (a rule base that does not yet exist), and the learned policy. When the two agree, the recommendation is consistent and explainable; when they disagree, either the rule has reached its limit or the policy is wrong, possibly informing future plan reviews. A large language model (LLM) based layer \citep{sano2026egu} can then help explain the recommendations and the disagreements, connect them to the emergency plan and available information, and support a human-in-the-loop decision process.

\makeatletter\if@neuripsfinal
\section*{Acknowledgments and Disclosure of Funding}
This work was supported by the European Union through the Marie Sk\l{}odowska-Curie Fellowship EXPEDITE, ``Exploring Opportunities for Developing a Risk and Resilience Climate Service Based on Big Data and Machine Learning'' (Grant Agreement No.~101067784). This work took place within the framework of the DoE 2023--2027 (MUR, AIS.DIP.ECCELLENZA2023\_27.FF project). The authors also acknowledge the Horizon 2020 MYRIAD-EU project (Grant Agreement No.~101003276) for its contribution to the broader development of multi-hazard and multi-risk research relevant to this study.
\fi\makeatother

\bibliographystyle{unsrtnat}
\bibliography{references}

@misc{schulman2017ppo,
  author        = {Schulman, John and Wolski, Filip and Dhariwal, Prafulla and Radford, Alec and Klimov, Oleg},
  title         = {Proximal Policy Optimization Algorithms},
  year          = {2017},
  eprint        = {1707.06347},
  archivePrefix = {arXiv},
  primaryClass  = {cs.LG},
  doi           = {10.48550/arXiv.1707.06347},
  url           = {https://arxiv.org/abs/1707.06347},
  note = {arXiv preprint arXiv:1707.06347}
}

@inproceedings{huang2020closer,
  author = {Huang, Shengyi and Onta{\~n}{\'o}n, Santiago},
  title = {A Closer Look at Invalid Action Masking in Policy Gradient Algorithms},
  booktitle = {Proceedings of the 35th International Florida Artificial Intelligence Research Society Conference ({FLAIRS}-35)},
  year = {2022},
  doi = {10.32473/flairs.v35i.130584},
  eprint = {2006.14171},
  archivePrefix = {arXiv},
  primaryClass = {cs.LG},
  url = {https://doi.org/10.32473/flairs.v35i.130584}
}

@inproceedings{kanervisto2020actionspace,
  author    = {Kanervisto, Anssi and Scheller, Christian and Hautam{\"a}ki, Ville},
  title     = {Action Space Shaping in Deep Reinforcement Learning},
  booktitle = {2020 IEEE Conference on Games (CoG)},
  publisher = {IEEE},
  year      = {2020},
  pages     = {479--486},
  doi       = {10.1109/CoG47356.2020.9231687},
  eprint    = {2004.00980},
  archivePrefix = {arXiv},
  primaryClass  = {cs.LG}
}

@inproceedings{velickovic2018gat,
  author    = {Veli{\v{c}}kovi{\'c}, Petar and Cucurull, Guillem and Casanova, Arantxa and Romero, Adriana and Li{\`o}, Pietro and Bengio, Yoshua},
  title     = {Graph Attention Networks},
  booktitle = {International Conference on Learning Representations (ICLR)},
  year      = {2018},
  url       = {https://openreview.net/forum?id=rJXMpikCZ},
  eprint    = {1710.10903},
  archivePrefix = {arXiv},
  primaryClass  = {stat.ML}
}

@article{feng2025rlcoastal,
  author  = {Feng, Kairui and Lin, Ning and Kopp, Robert E. and Xian, Siyuan and Oppenheimer, Michael},
  title   = {Reinforcement Learning--Based Adaptive Strategies for Climate Change Adaptation: An Application for Coastal Flood Risk Management},
  journal = {Proceedings of the National Academy of Sciences},
  volume  = {122},
  number  = {12},
  pages   = {e2402826122},
  year    = {2025},
  doi     = {10.1073/pnas.2402826122}
}

@article{shuvo2022multiagentslr,
  author  = {Shuvo, Salman Sadiq and Yilmaz, Yasin and Bush, Alan and Hafen, Mark},
  title   = {Modeling and Simulating Adaptation Strategies Against Sea-Level Rise Using Multiagent Deep Reinforcement Learning},
  journal = {IEEE Transactions on Computational Social Systems},
  volume  = {9},
  number  = {4},
  pages   = {1185--1196},
  year    = {2022},
  doi     = {10.1109/TCSS.2021.3122282}
}

@inproceedings{costa2024climaterl,
  author        = {Costa, Miguel and Petersen, Morten W. and Vandervoort, Arthur and Drews, Martin and Morrissey, Karyn and Pereira, Francisco C.},
  title         = {Climate Adaptation with Reinforcement Learning: Experiments with Flooding and Transportation in {Copenhagen}},
  booktitle     = {Tackling Climate Change with Machine Learning Workshop at NeurIPS 2024},
  year          = {2024},
  eprint        = {2409.18574},
  archivePrefix = {arXiv},
  primaryClass  = {cs.LG},
  url           = {https://www.climatechange.ai/papers/neurips2024/39}
}

@article{tang2024transitevac,
  author        = {Tang, Fang and Wang, Han and Delle Monache, Maria Laura},
  title         = {Strategizing Equitable Transit Evacuations: A Data-Driven Reinforcement Learning Approach},
  journal       = {Transportation Research Part C: Emerging Technologies},
  volume        = {180},
  pages         = {105342},
  year          = {2025},
  doi           = {10.1016/j.trc.2025.105342},
  eprint        = {2412.05777},
  archivePrefix = {arXiv}
}

@mastersthesis{viola2024evacrl,
  author = {Viola, Nicolas Pablo},
  title  = {A Reinforcement Learning Approach to Real-Time Emergency Evacuation},
  school = {Uppsala University, Department of Information Technology},
  year   = {2024},
  type   = {Master's thesis},
  url    = {https://urn.kb.se/resolve?urn=urn:nbn:se:uu:diva-536728},
  note = {Degree project 30 credits, Master Programme in Data Science, report IT mDA 24 012}
}

@article{dulacarnold2021realworld,
  author  = {Dulac-Arnold, Gabriel and Levine, Nir and Mankowitz, Daniel J. and Li, Jerry and Paduraru, Cosmin and Gowal, Sven and Hester, Todd},
  title   = {Challenges of Real-World Reinforcement Learning: Definitions, Benchmarks and Analysis},
  journal = {Machine Learning},
  volume  = {110},
  number  = {9},
  pages   = {2419--2468},
  year    = {2021},
  doi     = {10.1007/s10994-021-05961-4}
}

@article{ferrarin2020adriaticstorm,
  author = {Ferrarin, Christian and Valentini, Andrea and Vodopivec, Martin and Klaric, Dijana and Massaro, Giovanni and Bajo, Marco and De Pascalis, Francesca and Fadini, Amedeo and Ghezzo, Michol and Menegon, Stefano and Bressan, Lidia and Unguendoli, Silvia and Fettich, Anja and Jerman, Jure and Li{\v{c}}er, Matja{\v{z}} and Fustar, Lidija and Papa, Alvise and Carraro, Enrico},
  title = {Integrated sea storm management strategy: the 29 {October} 2018 event in the {Adriatic} {Sea}},
  journal = {Natural Hazards and Earth System Sciences},
  volume = {20},
  number = {1},
  pages = {73--93},
  year = {2020},
  doi = {10.5194/nhess-20-73-2020},
  url = {https://doi.org/10.5194/nhess-20-73-2020}
}

@article{lionello2021venicereview,
  author  = {Lionello, Piero and Nicholls, Robert J. and Umgiesser, Georg and Zanchettin, Davide},
  title   = {Venice Flooding and Sea Level: Past Evolution, Present Issues, and Future Projections (Introduction to the Special Issue)},
  journal = {Natural Hazards and Earth System Sciences},
  volume  = {21},
  number  = {8},
  pages   = {2633--2641},
  year    = {2021},
  doi     = {10.5194/nhess-21-2633-2021},
  url     = {https://doi.org/10.5194/nhess-21-2633-2021}
}

@article{gomezrave2025preparedness,
  author = {Gomez Rave, Dina Vanessa and Scolobig, Anna and del Jesus, Manuel},
  title = {Review article: {Rethinking} preparedness for coastal compound flooding: insights from a systematic review},
  journal = {Natural Hazards and Earth System Sciences},
  volume = {25},
  number = {10},
  pages = {3977--4002},
  year = {2025},
  doi = {10.5194/nhess-25-3977-2025},
  url = {https://doi.org/10.5194/nhess-25-3977-2025}
}

@article{dalbarco2025coastaid,
  author  = {Dal Barco, Maria Katherina and Casartelli, Veronica and San{\`o}, Marcello and Vascon, Sebastiano and Torresan, Silvia and Critto, Andrea},
  title   = {Prioritise Risks and Improve Adaptation Strategies in the {Veneto} Coast through the Application of a Custom {AI} Tool},
  journal = {International Journal of Disaster Risk Reduction},
  volume  = {130},
  pages   = {105818},
  year    = {2025},
  doi     = {10.1016/j.ijdrr.2025.105818}
}

@misc{osm,
  author       = {{OpenStreetMap contributors}},
  title        = {{OpenStreetMap} planet data},
  year         = {2024},
  howpublished = {\url{https://www.openstreetmap.org}},
  note         = {Planet dump retrieved from \url{https://planet.openstreetmap.org}; licensed under the Open Database License (ODbL) 1.0, \url{https://www.openstreetmap.org/copyright}}
}

@article{tinitaly,
  author  = {Tarquini, Simone and Isola, Ilaria and Favalli, Massimiliano and Mazzarini, Francesco and Bisson, Marina and Pareschi, Maria Teresa and Boschi, Enzo},
  title   = {{TINITALY/01}: a new Triangular Irregular Network of {Italy}},
  journal = {Annals of Geophysics},
  volume  = {50},
  number  = {3},
  pages   = {407--425},
  year    = {2007},
  doi     = {10.4401/ag-4424}
}

@inproceedings{sano2026egu,
  author    = {Sano, Marcello and Ferrario, Davide and Casagrande, Samuele and Vascon, Sebastiano and Torresan, Silvia and Critto, Andrea},
  title     = {Deep Reinforcement Learning for Operational Coastal Emergency Response With {AI} Agent Orchestration and Human Oversight},
  booktitle = {EGU General Assembly 2026},
  address   = {Vienna, Austria},
  month     = may,
  year      = {2026},
  note      = {Abstract EGU26-17369, session VPS14},
  doi       = {10.5194/egusphere-egu26-17369},
  url       = {https://meetingorganizer.copernicus.org/EGU26/EGU26-17369.html}
}

@article{bentivoglio2023swegnn,
  title   = {Rapid Spatio-Temporal Flood Modelling via Hydraulics-Based Graph Neural Networks},
  author  = {Bentivoglio, Roberto and Isufi, Elvin and Jonkman, Sebastiaan Nicolas and Taormina, Riccardo},
  journal = {Hydrology and Earth System Sciences},
  volume  = {27},
  number  = {23},
  pages   = {4227--4246},
  year    = {2023},
  doi     = {10.5194/hess-27-4227-2023},
  url     = {https://doi.org/10.5194/hess-27-4227-2023}
}

@inproceedings{meirom2021graphcontrol,
  author    = {Meirom, Eli A. and Maron, Haggai and Mannor, Shie and Chechik, Gal},
  title     = {Controlling Graph Dynamics with Reinforcement Learning and Graph Neural Networks},
  booktitle = {Proceedings of the 38th International Conference on Machine Learning (ICML)},
  series    = {Proceedings of Machine Learning Research},
  volume    = {139},
  pages     = {7565--7577},
  publisher = {PMLR},
  year      = {2021},
  url       = {https://proceedings.mlr.press/v139/meirom21a.html}
}

@inproceedings{yoon2021l2rpn,
  author    = {Yoon, Deunsol and Hong, Sunghoon and Lee, Byung-Jun and Kim, Kee-Eung},
  title     = {Winning the {L2RPN} Challenge: Power Grid Management via Semi-{M}arkov Afterstate Actor-Critic},
  booktitle = {International Conference on Learning Representations (ICLR)},
  year      = {2021},
  url       = {https://openreview.net/forum?id=LmUJqB1Cz8}
}

@misc{comunevenezia2008piano,
  author       = {{Comune di Venezia}},
  title        = {Piano Comunale di Protezione Civile},
  year         = {2009},
  howpublished = {Settore Protezione Civile, Deliberazione di Consiglio Comunale n.~15, 18 February 2009},
  url          = {https://www.comune.venezia.it/it/content/piano-comunale-protezione-civile}
}

\end{document}